\documentclass[11pt]{article}
\usepackage{mathptmx}

\usepackage[latin9]{inputenc}
\usepackage{geometry}
\usepackage{color}
\usepackage{array}
\usepackage{amsmath}
\usepackage{amssymb}
\usepackage{graphicx}
\makeatletter

\providecommand{\tabularnewline}{\\}

\usepackage{flushend}

\usepackage{graphicx} % more modern
\date{Feb 2015}

\makeatother

\begin{document}

\title{Learning deep representation of multityped objects and tasks}

\author{Truyen Tran, Dinh Phung and Svetha Venkatesh\\
Deakin University, Geelong, Australia}
\maketitle
\begin{abstract}
We introduce a deep multitask architecture to integrate multityped
representations of multimodal objects. This multitype exposition is
less abstract than the multimodal characterization, but more machine-friendly,
and thus is more precise to model. For example, an image can be described
by multiple visual views, which can be in the forms of bag-of-words
(counts) or color/texture histograms (real-valued). At the same time,
the image may have several social tags, which are best described using
a sparse binary vector. Our deep model takes as input multiple type-specific
features, narrows the cross-modality semantic gaps, learns cross-type
correlation, and produces a high-level homogeneous representation.
At the same time, the model supports heterogeneously typed tasks.
We demonstrate the capacity of the model on two applications: social
image retrieval and multiple concept prediction. The deep architecture
produces more compact representation, naturally integrates multiviews
and multimodalities, exploits better side information, and most importantly,
performs competitively against baselines. 
\end{abstract}

\global\long\def\hb{\boldsymbol{h}}
\global\long\def\h{h}
\global\long\def\vb{\boldsymbol{v}}
\global\long\def\v{v}
\global\long\def\wb{\boldsymbol{w}}
\global\long\def\yb{\boldsymbol{y}}
 \global\long\def\w{w}
\global\long\def\Vb{\boldsymbol{V}}
\global\long\def\Wb{\boldsymbol{W}}
\global\long\def\bb{\boldsymbol{b}}
\global\long\def\cb{\boldsymbol{c}}
\global\long\def\thetab{\boldsymbol{\theta}}

\section{Introduction\label{sec:intro}}

\begin{figure}
\begin{centering}
\begin{tabular}{>{\raggedright}p{2.1cm}>{\raggedright}p{0.1\textwidth}>{\raggedright}p{0.3\textwidth}>{\raggedright}p{0.15\textwidth}>{\raggedright}p{0.15\textwidth}}
\hline 
\textbf{\small{}Image} & \textbf{\small{}Caption} & \textbf{\small{}Tags} & \textbf{\small{}Comments} & \textbf{\small{}Groups}\tabularnewline
\hline 
{\small{}\vspace{1mm}
\includegraphics[width=2cm]{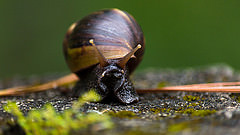}} & {\small{}\vspace{1mm}
i am moving} & {\small{}leaves, snail, slow, evening, movement, animal, crust, nikon,
d5100, srilanka, beautiful, art, asiasociety, image, photo, interesting,
green, mail, wait, ambition, colour, depth, bokeh} & {\small{}{[}amazing shot!, fabulous macro{]};}{\small \par}

{\small{}{[}awesome shot! what an amazing little creature!{]}} & {\small{}{[}insects, spiders, snails and slugs{]};}{\small \par}

{\small{}{[}fotógrafos anônimos{]}}\tabularnewline
\hline 
{\small{}\vspace{1mm}
\includegraphics[width=2cm]{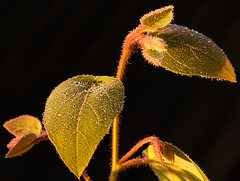}} & {\small{}\vspace{1mm}
leaves and dew} & {\small{}dew, nikon, d5100, srilanka, beautiful, art, asiasociety,
image, photo, interesting, morning, sun, sunlight, background, green} & {\small{}excellent light and beautiful tones against the black background,
uditha!!!} & {\small{}{[}fotógrafos anônimos{]}}\tabularnewline
\hline 
{\small{}\vspace{1mm}
\includegraphics[width=2cm]{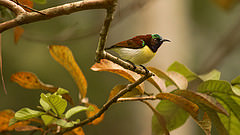}} & {\small{}\vspace{1mm}
purple-rumped sunbird} & {\small{}bird, nikon, d5100, srilanka, beautiful, art, asiasociety,
image, photo, interesting, photography, sunbird, purplerumpedsunbird,
branch, stance, wild, garden} & {\small{}a beautiful looking bird. well captured. congrats.} & {\small{}{[}birds{]}; {[}555{]}}\tabularnewline
\hline 
{\small{}\vspace{1mm}
\includegraphics[width=2cm]{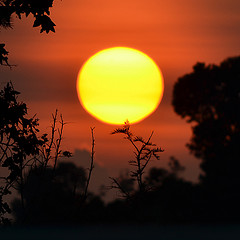}} & {\small{}\vspace{1mm}
wish you all happy 2015.} & {\small{}sunrise, newyear, nikon, d5100, srilanka, beautiful, art,
asiasociety, sun morning, leave, leaves, new, 2015, horizon, hospital,
sunspots, badulla, uva, year, rise, rising} & {\small{}wow! excellent shot! happy year!!!} & {\small{}{[}monochrome (color only!){]}; {[}fotógrafos anônimos{]}}\tabularnewline
\hline 
\end{tabular}
\par\end{centering}

\protect\caption{\label{fig:Flickr-images}Flickr images: captions, tags, comments
and groups. The visual modality can has multiple representations,
including a variety of histograms and bag of visual words. Photos
by \textbf{uditha wickramanayaka}.}
\end{figure}

Multimedia objects instantly command our attention. Yet they are notoriously
difficult to model. Multimedia objects require fusing multiple modalities
of distinct characteristics with wide semantic gaps. An online image,
for example, could be associated with human-level semantics such as
captions, tags, comments, sentiments and the surrounding context.
The image itself, however, has pixels and colors arranged in a 2D
grid -- a machine-level representation (see Fig.~\ref{fig:Flickr-images}
for examples). To construct more meaningful abstraction, a visual
image can be represented in multiple ways including several histograms
of color, edge, texture; and a bag of visual words (BOW) \cite{chua2009nus}. 

A common approach to learning from multiple representations is to
concatenate all features and form a long flat vector, assuming that
representations are equally informative. This is likely to cause problems
for several reasons. First there is a high variation in dimensionality
and scales between types (e.g., histogram vectors are often dense
and short while BOW vector is sparse and long). Second, modalities
have different correlation structures that warrant separate modeling
\cite{JMLR:v15:srivastava14b}. A single modality can also admit
multiviews, hence multiple types. For the visual modality, a histogram
is a real-valued vector but a bag of visual words is a discrete set
of word counts. Thus effective modeling should also account for type-specific
properties. Type modeling is, however, largely missing in prior multimodal
research \cite{barnard2003matching,ngiam2011multimodal,rasiwasia2010new}.
To better address multimodal differences in correlation structures,
it is necessary to recognize that intra-mode correlation is often
much higher than inter-mode correlation \cite{ngiam2011multimodal}.
This suggests a two step fusion. Translated to typed representations,
we need to first decorrelate the intra-type data, then capture the
inter-type inter-dependencies. 

While modeling multityped multimodality is essential to capture the
intra/inter-type correlation structure, care must be paid for the
tasks at hand. Representations that are handcrafted independent of
future tasks are likely to be suboptimal. This suggests \emph{task-informed}
representation learning. Putting together, we argue that a more successful
approach should work across tasks, modalities, semantic levels and
coding types. 

To this end, this paper introduces a \emph{bottom-up deep learning}
approach to tackle the challenges. Deep architectures \cite{hinton2006rdd,salakhutdinov2009deep_FULL,vincent2010stacked}
offer a natural solution for decoupling the intra-type correlation
from the inter-type correlation through multiple layers \cite{ngiam2011multimodal,JMLR:v15:srivastava14b}.
Starting from the bottom with machine-friendly representations, each
object is characterized by a collection of type-specific feature vectors,
where each type is a homogeneous numerical or qualitative representation
such as real-valued, binary, or count. These primitive representations
enable each modality to be expressed in more than one way, each of
which may belong to a type. For example, the visual modality of an
image can be represented by one or more \emph{real-valued} color histograms
or a bag of word \emph{counts}, and its social tags can be considered
as a sparse \emph{binary} vector. The intra-type correlation is handled
by type-specific lower models. Moving up the hierarchy, inter-type
correlation is captured in a higher layer. And finally, tasks are
defined on top of the higher layer.

\begin{figure*}
\begin{centering}
\begin{tabular}{>{\centering}p{0.23\textwidth}cc}
\includegraphics[width=0.2\textwidth]{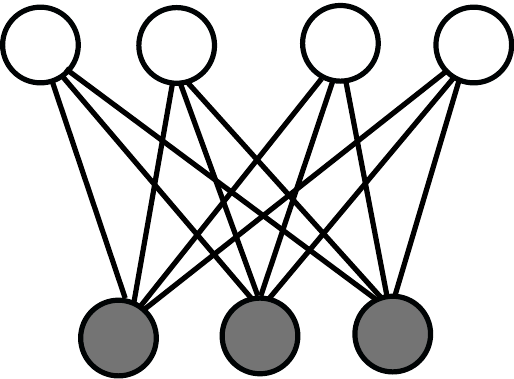} & \includegraphics[width=0.32\textwidth]{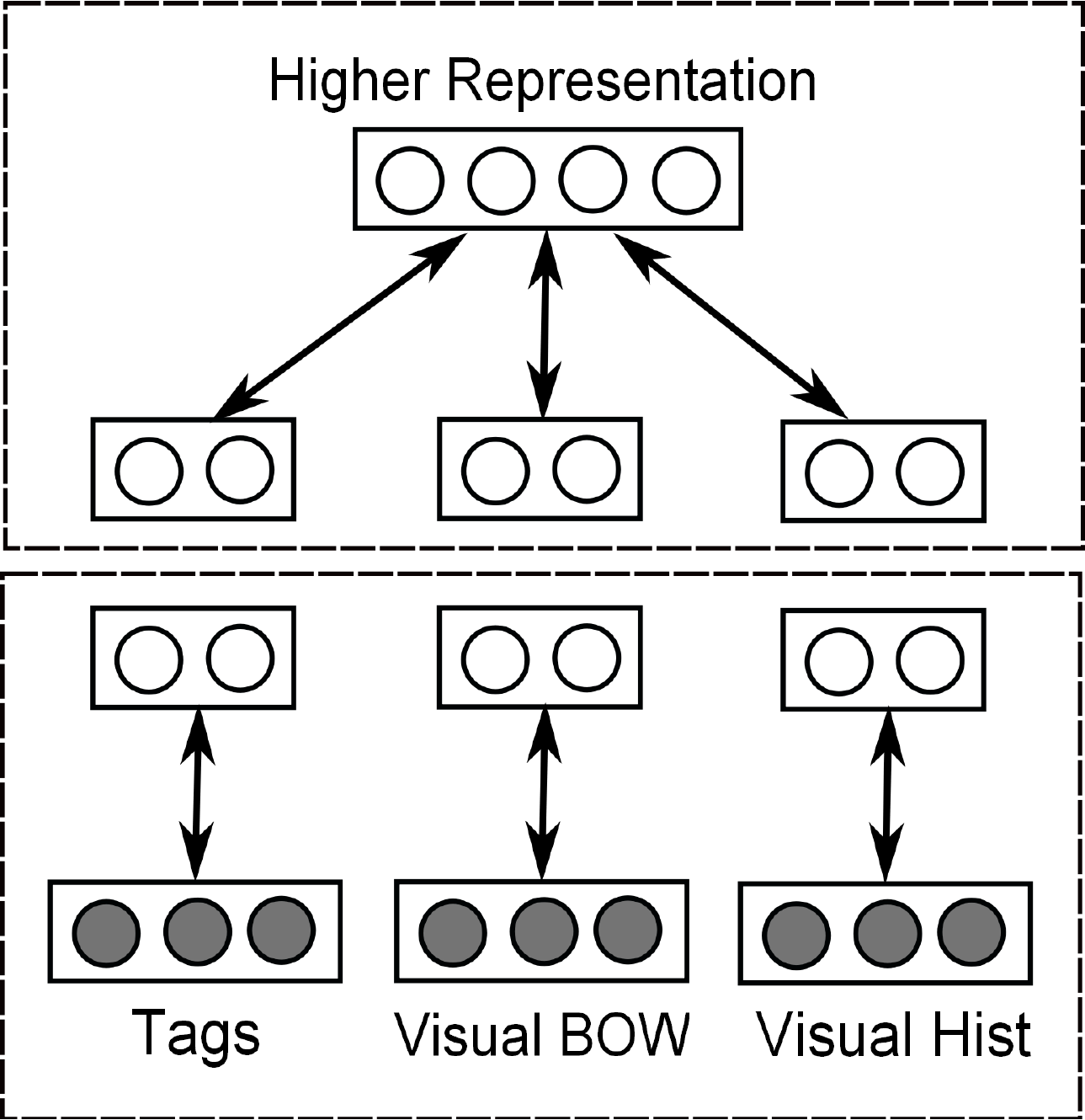} & \includegraphics[width=0.3\textwidth]{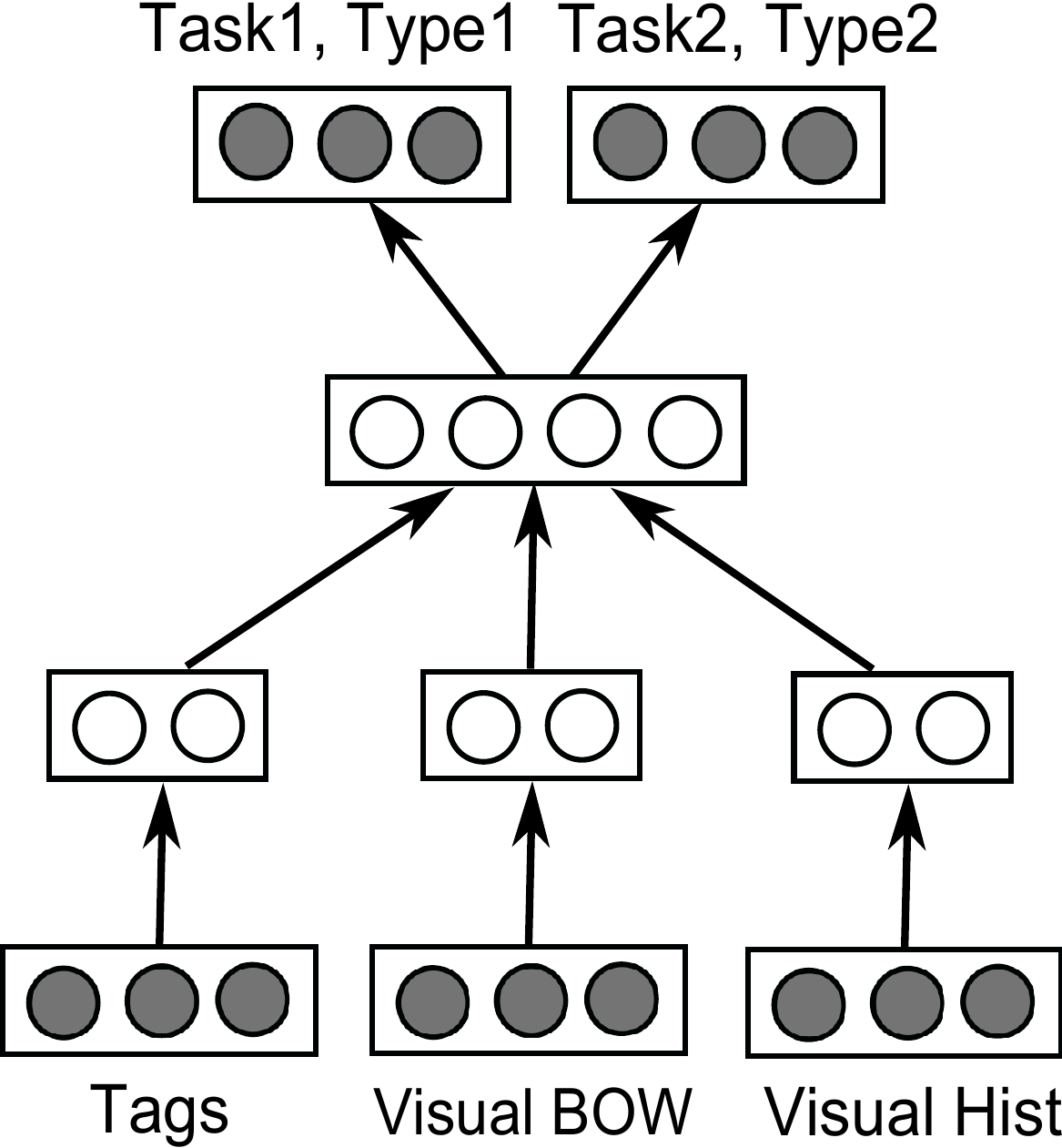}\tabularnewline
(A) Restricted Boltzmann machine & (B) Layerwise training & (C) Multitask multityped refinement\tabularnewline
\end{tabular}
\par\end{centering}

\protect\caption{A $3$-step training procedure: (A) First type-specific RBMs are trained;
(B) then posteriors of the previous layers are used as the input for
the next layer; (C) finally auxiliary tasks are learnt, refining the
previously discovered latent representation. Steps A, B are unsupervised
and Step C is supervised. \label{fig:Models}}
\end{figure*}

Our realization of this deep learning scheme is a two-phase procedure
based on a stack of restricted Boltzmann machines (RBMs) \cite{smolensky1986information,welling2005efh_FULL},
as depicted in Fig.~\ref{fig:Models}. The first learning phase is
unsupervised. It discovers data regularities and the shared structures
between modalities and types without labels. Learning starts from
the lowest level, where an array of RBMs map the type-specific inputs
into intermediate representations. These middle-level representations
are then integrated by another binary RBM into high-level features.
The stack forms a multityped deep belief network \cite{hinton2006rdd}.
In the second learning phase, auxiliary typed tasks are introduced
on top of the highest feature layer, essentially forming a multityped
deep neural network. The tasks are learnt simultaneously and the previously
learnt representations are refined in the process. This creates a
more predictive deep representation shared among tasks \cite{bengio2013representation}.
Our approach appears to resemble recent multimodal multitask learning
frameworks \cite{zhang2012multi,xie2014cross}. However, we differ
significantly by modeling types at both the input and the task levels,
while previous work often assumes homogeneous representations.

To summarize, this paper makes the following contributions:
\begin{itemize}
\item Introduction of a deep multityped architecture to integrate multiviews
and multimodalities.
\item Introducing the concept of multityped multitask learning, where tasks
are heterogeneous in types.
\item An evaluation of the proposed architecture on the NUS-WIDE data \cite{chua2009nus},
demonstrating its capacity on two applications: image retrieval and
multilabel concept learning.
\end{itemize}
The next section reviews related background. Section~\ref{sec:Restricted-Boltzmann-machines}
presents the building blocks of the proposed architecture -- restricted
Boltzmann machines. The deep architecture for multityped inputs and
tasks is described in Section~\ref{sec:A-deep-architecture}. Section~\ref{sec:Experiments}
demonstrates the applicability of the proposed deep architecture on
two applications: image retrieval and multilabel learning. Finally,
Section~\ref{sec:Conclusion} concludes the paper.

\section{Related background\label{sec:bg}}

Learning from multimodal data has recently attracted an increasing
attention \cite{ngiam2011multimodal,rasiwasia2010new,duchnowski1994see,guillaumin2010multimodal,ozdemir2014probabilistic,yuhas1989integration}.
However, the fusion of multiple data types is sparsely studied. Early
attempts include the work of \cite{barnard2003matching}, where a
joint probabilistic model, the correspondence latent Dirichlet allocation
(LDA), was proposed. The model explicitly assumes the direct correspondence
in the data (i.e., which image region generates which labels). A more
recent work \cite{rasiwasia2010new} uses LDA to model text and image
features. Modalities are treated as two separate sources which are
lately fused using canonical correlation analysis. It is, however,
not clear to how systematically handle different data types and more
than two sources.

Our work extends a rich class of undirected graphical models known
as restricted Boltzmann machines (RBMs) \cite{smolensky1986information},
capable of encoding arbitrary types and side information \cite{Truyen:2011b,truyen_phung_venkatesh_icml13}.
Most existing work in RBMs is limited to single types, for example,
binary \cite{smolensky1986information}, real-valued \cite{hinton2006rdd}
or count \cite{salakhutdinov2009semantic}. Joint modelling of visual
features and tags was first introduced in \cite{xing2005mining}.
This line of work was extended further in \cite{Truyen:2011b,truyen_phung_venkatesh_icml13}
with a richer set of types. These works, however, assume a shallow
architecture which may not be ideal for multimedia objects, where
modalities differ substantially in their correlation structures and
semantic levels \cite{ngiam2011multimodal}.

More recently, deep architectures have been suggested to overcome
the difficulties associated with the shallow models in multimodal
learning. The deep canonical correlation analysis \cite{andrew2013deep},
for example, can potentially help narrow the semantic gaps. Multimodal
stacked autoencoders have been shown to be effective for integrating
speech and vision \cite{ngiam2011multimodal}. These efforts, however,
do not address multiple data types. Closer to our architecture is
perhaps the work of \cite{JMLR:v15:srivastava14b}, where a deep
Boltzmann machine \cite{salakhutdinov2009deep_FULL} is used for
multimodal data. The key differences are that the work of \cite{JMLR:v15:srivastava14b}
assumes each modality has only one view (and type), and the model
is purely generative. Our work, on the other hand, extends to multiple
views and multiple types per modality, with both the generative and
discriminative components. Multiple views (and types) from the same
visual modality has been studied in \cite{kang2012deep}, but this
is limited to hashing and does not consider multityped tasks.

Neural networks for multitask learning was first introduced in \cite{caruana1997multitask}.
Exploiting auxiliary tasks with shared latent representation to improve
the main task has been previously proposed \cite{ando2005framework}.
Multimodal multitask learning has been studied in \cite{zhang2012multi,xie2014cross}.
However, we differ significantly by modeling heterogeneous types at
both the input and task levels, while the previous work often assumes
homogeneous representations.

\section{Sparse restricted Boltzmann machines \label{sec:Restricted-Boltzmann-machines}}

We first present building blocks for our deep architecture -- the
restricted Boltzmann machine (RBM) and its generalizations. A restricted
Boltzmann machine is a stochastic undirected neural network without
outputs \cite{smolensky1986information,welling2005efh_FULL} (see
Fig.~\ref{fig:Models}(A) for a graphical depict). Given binary inputs
$\vb\in\left\{ 0,1\right\} ^{N}$ and hidden units $\hb\in\left\{ 0,1\right\} ^{K}$,
the RBM defines the following energy function:
\begin{equation}
E(\vb,\hb)=-\sum_{i=1}^{N}a_{i}v_{i}-\sum_{k=1}^{K}b_{k}h_{k}-\sum_{i=1}^{N}\sum_{k=1}^{K}W_{ik}v_{i}h_{k}\label{eq:binary-energy}
\end{equation}
where $a_{i},b_{k}$ are biases and $W_{ik}$ is the weight of the
interaction between an input unit $i$ and a hidden unit $k$. 

Unlike traditional neural networks, the RBM does not map inputs into
outputs, but rather, models a joint multivariate distribution as follows:
\[
P(\vb,\hb)=\frac{1}{Z}\exp\left(-E(\vb,\hb)\right)
\]
where $Z$ is the normalizing constant. The RBM's bipartite structure
(Fig.~\ref{fig:Models}a) posits the following factorizations
\begin{eqnarray}
P(\vb\mid\hb) & = & \prod_{i=1}^{N}P(v_{i}\mid\hb);\quad P(\hb\mid\vb)=\prod_{k=1}^{K}P(h_{k}\mid\vb)\label{eq:factorizations}
\end{eqnarray}
where
\begin{align}
P(v_{i} & =1\mid\hb)=\sigma\left[a_{i}+\sum_{k=1}^{K}W_{ik}h_{k}\right]\label{eq:generative-1}\\
P(h_{k} & =1\mid\vb)=\sigma\left[b_{k}+\sum_{i=1}^{N}W_{ik}v_{i}\right]\label{eq:posterior}
\end{align}
where $\sigma[z]=1/(1+e^{-z})$ is the sigmoid function. The posterior
vector $\left(\bar{h}_{1},\bar{h}_{2},...,\bar{h}_{K}\right)$ where
$\bar{h}_{k}=P(h_{k}=1\mid\vb)$ serves as a new representation of
the data. The factorizations in Eq.~(\ref{eq:factorizations}) enable
efficient layerwise MCMC sampling by alternating between $\hat{\vb}\sim P(\vb\mid\hb)$
and $\hat{\hb}\sim P(\hb\mid\vb)$.

Parameters $\thetab=\left\{ a_{i},b_{k},W_{ik}\right\} $ are typically
estimated by maximizing the data likelihood $P(\vb)$. To encourage
\emph{sparse representation}, that is, the low probability that a
hidden unit is activated, we augment the data log-likelihood with
a regularization term:
\[
\thetab^{*}=\arg\max_{\thetab}\left(\log P(\vb)+\gamma\sum_{k}q_{k}\bar{h}_{k}\right)
\]
where $q_{k}\ll1$ is the desired activation probability of the $k$-th
hidden unit, and $\gamma>0$ is regularization factor. Learning can
be realized through a fast stochastic gradient ascent procedure known
as contrastive divergence \cite{Hinton02}:
\begin{align}
W_{ik} & \leftarrow W_{ik}+\eta\left(v_{i}\bar{h}_{k}-\bar{v}_{i}\hat{h}_{k}\right)+\gamma v_{i}\left(q_{k}-\bar{h}_{k}\right)\label{eq:param-update-map}\\
b_{k} & \leftarrow b_{k}+\eta\left(\bar{h}_{k}-\hat{h}_{k}\right)+\gamma\left(q_{k}-\bar{h}_{k}\right)\label{eq:param-update-hidden}\\
a_{i} & \leftarrow a_{i}+\eta\left(v_{i}-\bar{v}_{i}\right)\label{eq:param-update-visible}
\end{align}
where $\eta>0$ is learning rate, $\bar{v}_{i}=P(v_{i}=1\mid\hat{\hb})$,
and $\hat{\hb}$ is a sample drawn from a short MCMC run starting
from the input $\vb$. This procedure approximately maximizes the
data likelihood $P(\vb)$.

A major limitation of RBM is that it can only model binary inputs.
In the next subsections, we present extensions for real-valued inputs
with Gaussian-Bernoulli RBM \cite{hinton2006rdd} and count inputs
with constrained Poisson RBM \cite{salakhutdinov2009semantic}.

\subsection{Gaussian--Bernoulli RBM}

Gaussian--Bernoulli RBMs model real--valued inputs such as pixel intensities
and histograms \cite{hinton2006rdd}. The graphical structure is
exactly the same that of standard RBMs (Fig.~\ref{fig:Models}(A)),
except for the input type, where $\vb\in\mathbb{R}^{N}$. The energy
in Eq.~(\ref{eq:binary-energy}) is replaced by:
\[
E(\vb,\hb)=\frac{1}{2}\sum_{i=1}^{N}\left(v_{i}-a_{i}\right)^{2}-\sum_{k=1}^{K}b_{k}h_{k}-\sum_{i=1}^{N}\sum_{k=1}^{K}W_{ik}v_{i}h_{k}
\]
Here for simplicity, we have assumed unit variance for each input
variable. Like the binary case, the factorizations in Eq.~(\ref{eq:factorizations})
still hold. The posterior $P(h_{k}\mid\vb)$ assumes the same form
as in Eq.~(\ref{eq:posterior}). However, the generative distribution
$P(\v_{i}\mid\hb)$ now follows a normal distribution of variance
$1$ and mean:
\begin{equation}
\mu_{i}(\hb)=a_{i}+\sum_{k=1}^{K}W_{ik}h_{k}\label{eq:mean}
\end{equation}

Learning in the Gaussian--Bernoulli RBMs can also be done using the
stochastic procedure as in Eqs.~(\ref{eq:param-update-map}--\ref{eq:param-update-visible})
with a modification: $\bar{v}_{i}=\mu_{i}(\hb)$.

\subsection{Constrained Poisson RBM}

Counts as in bag-of-words representation can be modeled using an approximate
constrained Poisson model \cite{salakhutdinov2009semantic}. Let
$\vb\in\left\{ 0,1,2,...\right\} ^{N}$, the model energy reads
\[
E(\vb,\hb)=\sum_{i=1}^{N}\left(\log\left(v_{i}!\right)-a_{i}v_{i}\right)-\sum_{k=1}^{K}b_{k}h_{k}-\sum_{i=1}^{N}\sum_{k=1}^{K}W_{ik}v_{i}h_{k}
\]
The factorizations in Eq.~(\ref{eq:factorizations}) still hold and
does the posterior $P(h_{k}\mid\vb)$ in Eq.~(\ref{eq:posterior}).
However, the generative distribution $P(\v_{i}\mid\hb)$ now admits
a constrained Poisson distribution:
\[
P(\v_{i}=n\mid\hb)=\mbox{Poisson}\left(n,\lambda_{i}(\hb)\right)
\]
where $\lambda_{i}(\hb)$ is the mean--rate, calculated as
\[
\lambda_{i}(\hb)=M\frac{\exp\left(\mu_{i}(\hb)\right)}{\sum_{j=1}^{N}\exp\left(\mu_{j}(\hb)\right)}
\]
where $M=\sum_{i}v_{i}$ is document length and $\mu_{i}(\hb)$ is
given in Eq.~(\ref{eq:mean}). This special mean--rate ensures that
$\sum_{i}\lambda_{i}(\hb)=M$, i.e., $\lambda_{i}(\hb)$ is bounded
from above. In addition, the model rate equals the empirical rate,
on average, leading to more stable learning.

Learning in the constrained Poisson RBMs using the stochastic procedure
as in Eqs.~(\ref{eq:param-update-map}--\ref{eq:param-update-visible})
requires only a minimum change: $\bar{v}_{i}=\lambda_{i}(\hb)$.

\subsection{Other RBM types}

RBMs have been extended to other types including multinomial \cite{Salakhutdinov-et-alICML07_FULL},
ordinal \cite{Truyen:2009a}, rank \cite{Truyen:2011b}, beta \cite{le2011learning}
and a mixture of these \cite{Truyen:2011b}. As with real-valued
and count types, these extensions maintain the same bipartite structure
of the RBMs, only differ in type-specific generative distributions
$P(\v_{i}\mid\hb)$. A recent flexible framework in \cite{truyen_phung_venkatesh_icml13}
is claimed to represent most known types using truncated Gaussian--Bernoulli
RBMs. Here there are two hidden layers rather than one and the real-valued
layer is constrained through type-specific inequalities.

\section{A deep architecture for multityped representations and tasks \label{sec:A-deep-architecture}}

We now present the deep architecture for multityped objects and multityped
tasks. Our learning is a $3$-step procedure spanning over an unsupervised
learning phase and a supervised phase, as depicted in Figs.~\ref{fig:Models}
(A), (B) and (C), respectively. The key ideas are two-fold: (i) separation
of modeling of \emph{intra-type correlation} from \emph{inter-type
correlation} using multiple layers, and (ii) using predictive tasks
to refine the learnt representation.

\subsection{Learning procedure}

We derive a 2-phase learning procedure. The first phase is unsupervised
where representations are learnt from the data without guidance from
any labels. The second phase is supervised where the tasks are used
to fine-tune the previously learnt model.

\subsubsection*{Phase 1: Unsupervised learning}

This phase has two steps, one is to model intra-type correlation and
the other is to decouple inter-type correlation.
\begin{itemize}
\item \emph{Step A -- Learning type-specific representation}s: We first
ignore the inter-type correlation and each feature set is modeled
separately using a type-specific RBM. Let us assume that each object
can be represented by $S$ ways, each of which corresponds to a feature
set $\vb_{s}$, for $s=1,2,..,S$. At the end of the step, we have
$S$ set of posteriors $\hat{\hb}_{s}=\left(\hat{h}_{s1},\hat{h}_{s2},..,\hat{h}_{sN_{s}}\right)$,
where $\hat{h}_{sk}=P(h_{sk}=1\mid\vb_{s})$ for $s=1,2,...,S$. This
new representation is expected to reduce the intra-type correlation.
\item \emph{Step B -- Learning joint representation}: We now model the inter-type
correlation (e.g., see the upper half of Fig.~\ref{fig:Models}(B)).
The posteriors of type-specific RBMs for each object are concatenated
as an input for another binary RBM\footnote{Since we use the probabilities instead of binary values as input,
this is really an approximation.}, i.e., $\vb^{(2)}\leftarrow\left[\hat{\hb}_{1}^{(1)}\mid\hat{\hb}_{2}^{(1)}\mid...\mid\hat{\hb}_{S}^{(1)}\right]$.
The posterior in this second stage $P(\hb^{(2)}\mid\vb^{(2)})$ is
expected to be more abstract than those in the first stage. This step
has been shown to improve the lower-bound of the data log-likelihood:
$P(\vb)$ \cite{hinton2006rdd}. 
\end{itemize}

\subsubsection*{Phase 2: Supervised fine-tuning}
\begin{itemize}
\item \emph{Step (C) -- Learning tasks-informed representations}: The two
layers from Steps A and B are now connected by joining the output
of the first layer and the input of the second layer, as depicted
in Fig.~\ref{fig:Models}(C). Essentially, the joint model becomes
a generalization of the deep belief network \cite{hinton2006rdd}
where the input is multityped. Further, the top hidden layer is connected
to tasks of interest. The whole model is now a complex deep neural
network which supports multiple input types as well as multiple output
tasks. Finally, the parameters are refined by discriminative training
of tasks through back-propagation. We elaborate this step further
in the following subsection.
\end{itemize}

\subsection{Multityped multitask learning}

Let $\bb_{s}^{(1)},\Wb_{s}^{(1)}$ denote the parameters associated
with the $s$-th feature set at the bottom layer, and $\bb^{(2)},\Wb^{(2)}$
be the parameters of the second layer. At the end of Step B, the $k$-th
higher feature can be computed in a feedforward procedure:
\begin{eqnarray}
f_{k}(\vb_{1:S}) & = & \sigma\left[b_{k}^{(2)}+\sum_{s}\sum_{m}W_{smk}^{(2)}\sigma\left[b_{sm}^{(1)}+\sum_{i}W_{sim}^{(1)}v_{si}\right]\right]\label{eq:top-rep}
\end{eqnarray}
where $\sigma$ is the sigmoid function. These learned representations
can then be used for future tasks such as retrieval, clustering or
classification.

However, learning up to this point has been aimed to approximately
optimize the lower-bound of the data likelihood $P(\vb_{1:S})$ \cite{hinton2006rdd}.
Since the data likelihood may not be well connected with the tasks
we try to perform, using the discovered representation at the second
stage may be suboptimal. A more informative way is to guide the learning
process so that the learned representation become \emph{predictive}
with respect to future tasks. Beside the main tasks of interest, we
argue that auxiliary tasks can be helpful the shape the representations.
This provides a realization of Vapnik's idea of learning with privileged
information \cite{vapnik2009new} in that the teacher knows extra
information (auxiliary tasks) that is only available at training time.
For example, if the main task is content-based retrieval or concept
labeling, ranking social tags associated with images can be used as
an auxiliary task. On absence of well-defined extra tasks, we can
use input reconstruction as auxiliary tasks. Note that the auxiliary
tasks (and side information) are only needed at training time.

The major distinctive aspect of our multitask setting is that tasks
are multityped, that is, tasks do not assume the same functional form.
For example, one task can be histogram reconstructions through regression,
the other can be tagging through label ranking. The tasks, however,
share the same latent data representation, along the line of \cite{caruana1997multitask}.
As the model is now a deep neural network, the standard back-propagation
applies. The major difference is now the parameters learnt from the
unsupervised phase serve as initialization for the supervised phase.
Numerous evidences in the literature suggest that this initialization
is in a sensible parameter region, enabling exploitation of good local
minima, and at the same time, implicitly provides regularization \cite{hinton2006rdd,erhan2010does}.

We consider two task categories, one with a single outcome per task
(unstructured outputs), and the other with composite outcome (structured
outputs).

\subsubsection{Unstructured outputs \label{sub:Unstructured-outputs}}

Let $y_{t}$ be the outcome for the $t$-th task and 
\[
g_{t}(\vb_{1:S})=V_{t0}+\sum_{k}V_{tk}f_{k}(\vb_{1:S})
\]
where $f_{k}(\vb_{1:S})$ are top-level features computed in Eq.~(\ref{eq:top-rep}).
We then refine the model by optimizing loss functions with respect
to all parameters $\cb,\Vb,\Wb^{(1)},\Wb^{(2)},\bb_{1:S}^{(1)}$ and
$\bb^{(2)}$. Examples of task types and associated loss functions
are:
\begin{itemize}
\item \emph{Regression for real-valued outcomes}. For $y_{t}\in\mathbb{R}$,
the loss function is:
\begin{equation}
L_{t}=\frac{1}{2}\left(y_{t}-g_{t}(\vb_{1:S})\right)^{2}\label{eq:regression}
\end{equation}

\item \emph{Logistic regression for binary outcomes}. For $y_{t}\in\pm1$,
the loss function has a log-form:
\begin{equation}
L_{t}=\log\left(1+\exp\left(-y_{t}g_{t}(\vb_{1:S})\right)\right)\label{eq:logistic-regression}
\end{equation}

\item \emph{Poisson regression for counts}. For $y_{t}=0,1,2,..$, the loss
function is negative log of the Poisson distribution:
\begin{equation}
L_{t}=\log(y_{t}!)-y_{t}\log\lambda_{t}+\lambda_{t}\label{eq:poisson-regression}
\end{equation}
where $\lambda_{t}=e^{g_{t}(\vb_{1:S})}$ is the mean-rate. Since
the first term $\log(y_{t}!)$ is a constant with respect to model
parameters, it can be ignored during learning.
\end{itemize}

\subsubsection{Structured outputs \label{sub:Structured-outputs}}

We consider three cases, namely multiclass prediction, (partial) label
ranking and multilabel learning. Label ranking refers to the ordering
of labels according to their relevance with respect to an object.
For example, for an image, the ranking of the groups to which the
image belong may be: \texttt{\small{}\{birds} $\succ$ \texttt{\small{}close-up}
$\succ$ \texttt{\small{}outdoor} $\succ$ \texttt{\small{}indoor\}}.
Partial label ranking is when the rank given is incomplete, e.g.,
only the \texttt{\small{}\{birds} $\succ$ \texttt{\small{}close-up}
$\succ$ \texttt{\small{}others\}} is given. When only one label is
known, this reduces to the multiclass problem. Multilabel prediction
is to assign more than one label for each object. The same above image
can be tagged as \texttt{\small{}\{bird, morning, branch, wild, garden}{\small{}\},
for example.}{\small \par}

Let $\left\{ y_{t1},y_{t2},...,y_{tT_{t}}\right\} $ be the $T_{t}$
labels given to the object in the $t$-th task. Here we employ a simple
strategy: A label is chosen with the following probability:

\begin{equation}
P_{j}\left(y_{tj}=l\mid\vb_{1:S}\right)=\frac{\exp\left\{ V_{tl}+\sum_{k}V_{tlk}f_{k}(\vb_{1:S})\right\} }{\sum_{l'\in\mathcal{L}_{j}}\exp\left\{ V_{tl'}+\sum_{k}V_{tl'k}f_{k}(\vb_{1:S})\right\} }\label{eq:multinomial}
\end{equation}
where $\mathcal{L}_{j}$ is a set consisting of $y_{tj}$ and other
labels that are deemed equal or better than the $j$-th outcome. The
loss function is
\[
L_{t}=-\sum_{j=1}^{T_{t}}\log P_{j}\left(y_{tj}=l\mid\vb_{1:S}\right)
\]
The three specific cases are: 
\begin{itemize}
\item For \emph{multiclass prediction}, $\mathcal{L}_{1}$ has all class
labels and $T_{t}=1$.
\item For \emph{(partial) label ranking}, $\mathcal{L}_{1}$ consists of
all labels and $\mathcal{L}_{j}$ is defined recursively as follows
$\mathcal{L}_{j}=\mathcal{L}_{j-1}\backslash y_{tj}$. This essentially
reduces to the Plackett-Luce model when complete ranking is available
\cite{cheng2010label}. Prediction is simple as we need to rank all
the labels according to $P_{j}\left(y_{tj}=l\mid\vb_{1:S}\right)$. 
\item For \emph{multilabel learning}, $\mathcal{L}_{j}$ contains all labels
for $j=1,2,..,T_{t}$. For prediction, we adapt the strategy in \cite{elisseeff2001kernel}
in that only those labels satisfying $P(y\mid\vb_{1:S})\ge\tau(\vb_{1:S})$
are selected. The threshold $\tau(\vb_{1:S})$ is a function of $\left\{ P_{j}\left(y_{tj}=l\mid\vb_{1:S}\right)\right\} _{j=1}^{T_{t}}$
which is estimated from the training data by maximizing an objective
of interest. In our setting, the objective is the label-wise $F$-score.
\end{itemize}

\section{Experiments \label{sec:Experiments}}

We evaluate our learning method on the NUS-WIDE\footnote{http://lms.comp.nus.edu.sg/research/NUS-WIDE.htm}
dataset \cite{chua2009nus}. This consists of more than $200K$ Flickr
images, each of which is equipped with (1) $6$ visual feature representations
(64-D color histogram, 144-D color correlogram, 73-D edge direction
histogram, 128-D wavelet texture, 225-D block-wise color moments,
and 500-D BOWs from SIFT descriptors), (2) tags drawn from a $5$K-word
vocabulary, and (3) one or more of $81$ manually labelled concepts.
We randomly pick $10,000$ images for training our model and $10,000$
for testing. The dimensionality of the social tags is limited to $1000$.
For BOWs, we use $\left(\left[\log(1+\mbox{count})\right]\right)$
instead of\emph{ }count\emph{. }For real-valued histograms, we first
normalize each histogram to an unit vector, then normalize them across
all train data to obtain zero mean and unit variance.

Our model architecture has $200$ hidden units per type at the bottom
layer, and $200$ hidden units at the top layer. Mapping parameters
$\Wb$ are randomly initialized from small normally distributed numbers,
and bias parameters $\bb$ are from zeros. Posterior sparsity $q_{k}$
is set to $0.2$ (see Eq.~(\ref{eq:param-update-map})). Learning
rates are set at $0.1$ for binary, $0.01$ for real-valued and $0.02$
for count types. The difference in learning rate scales is due to
the fact that real-valued Gaussian means are unbounded, and Poisson
mean--rates, although bounded, may be larger than 1. Parameters in
the unsupervised phase (Steps A,B) are updated after every $100$
training images. The discriminative training in the supervised phase
(Step C) is based on back-propagation and conjugate gradients.

\subsection{Image retrieval \label{sub:Image-Retrieval}}

In this task, each test image is used to query other test objects.
Retrieval is based on the cosine similarity between representations
of the query and the objects. For our deep architecture, the top-level
representation is used, where each unit is $f_{k}(\vb_{1:S})$ as
in Eq.~(\ref{eq:top-rep}). To create a baseline representation,
we first normalize each feature set to a unit vector before concatenating
them. This eliminates the difference in dimensionality and scales
between views. A retrieved image is considered relevant if it shares
at least one manually labelled concept with the query.

\begin{table*}
\begin{centering}
\begin{tabular}{|l|l|l|}
\hline 
\textcolor{black}{\emph{Features}}\textcolor{black}{$\rightarrow$
}\textcolor{black}{\emph{Auxiliary Tasks}} & \textcolor{black}{\emph{Input types}} & \textcolor{black}{\emph{Output types}}\tabularnewline
\hline 
\textcolor{black}{500D-BOW$\rightarrow$ Tags} & \textcolor{black}{Count} & Multilabel\tabularnewline
\textcolor{black}{64D-CH$\rightarrow$ Tags } & \textcolor{black}{Real} & Multilabel\tabularnewline
\textcolor{black}{1134D-Visual$\rightarrow$Tags} & \textcolor{black}{Real \& count} & Multilabel\tabularnewline
\textcolor{black}{1134D-Visual$\rightarrow$(Visual \& Tags)} & \textcolor{black}{Real \& count} & \textcolor{black}{Real, count \& }multilabel\tabularnewline
\textcolor{black}{(64D-CH \& Tags)$\rightarrow$ Tags} & \textcolor{black}{Real \& bin} & Multilabel\tabularnewline
\hline 
\end{tabular}
\par\end{centering}

\protect\caption{Type definitions for retrieval. \textcolor{black}{(Features}\textcolor{black}{\emph{$\rightarrow$
}}\textcolor{black}{Aux. Tasks) means auxiliary tasks are used to
fine-tune the model given the features. 64D-CH means the 64 bin color
histogram, 1134D-Visual means all 6 visual features combined.} \label{tab:IR-types}}
\end{table*}

Auxiliary tasks are used in the training phase but not in the testing
phase. If the input is purely visual, then this reduces to \emph{content-based
image retrieval} \cite{smeulders2000content}. We test two settings
with the auxiliary tasks, where we predict either tags (tagging) or
combination of visual features and tags (feature reconstruction \&
tagging). The task types are defined as follows (see also Tab.~\ref{tab:IR-types}):
\begin{itemize}
\item \emph{Tagging} as a multilabel prediction problem (Sec.~\ref{sub:Structured-outputs}).
\item \emph{Histogram reconstruction} as multiple regressions (Eq.~(\ref{eq:regression})).
\item \emph{BOW reconstruction} as multiple Poisson regressions (Eq.~(\ref{eq:poisson-regression})).
\end{itemize}

\subsubsection{Performance measures}

Two performance measures are reported: Mean Average Precision (MAP)
over the top $100$ retrieved images, and the Normalized Discounted
Cumulative Gain (NDCG) \cite{jarvelin2002cumulated} at the top $10$
images. Let $\mbox{rel}_{i}\in\left\{ 0,1\right\} $ denote whether
the $i$-th retrieved image is relevant to the query. The precision
at rank $n$ is $\mbox{Precision}(n)=\frac{1}{n}\sum_{i=1}^{n}\mbox{rel}_{i}$.
The MAP is computed as follows:
\begin{align*}
\mbox{MAP} & =\frac{1}{Q}\sum_{q=1}^{Q}\frac{1}{T}\sum_{n=1}^{T}\mbox{Precision}_{q}(n)
\end{align*}
where $Q=10,000$ is the number of test queries, and $T=100$ is the
number of top images. 

The Discounted Cumulative Gain (DCG) for the top $T$ images is computed
as
\[
\mbox{DCG}@T=\sum_{n=1}^{T}\frac{\mbox{rel}_{n}}{\log_{2}(n+1)}
\]
This attains the maximum score if all $T$ images are relevant, i.e.,
$\mbox{max}\mbox{DCG}@T=\sum_{n=1}^{T}\frac{1}{\log_{2}(n+1)}$. Finally,
the NDCG score is computed as follows:
\begin{align*}
\mbox{NDCG}@T & =\frac{1}{Q}\sum_{q=1}^{Q}\frac{\mbox{DCG}_{q}@T}{\mbox{max}\mbox{DCG}_{q}@T}
\end{align*}
for $T=10$.

\subsubsection{Results}

\begin{table*}
\begin{centering}
\begin{tabular}{|l|l|l|c|c|c|r|}
\hline 
 & \multicolumn{3}{c|}{\textbf{Our method}} & \multicolumn{3}{c|}{\textbf{Baseline}}\tabularnewline
\cline{2-7} 
\textcolor{black}{\emph{Features}}\textcolor{black}{$\rightarrow$
}\textcolor{black}{\emph{Aux. Tasks}} & \textcolor{black}{\emph{MAP }}\textcolor{black}{($\uparrow$\%)} & \textcolor{black}{\emph{N@10}}\textcolor{black}{($\uparrow$\%)} & \textcolor{black}{\emph{Dim.}} & \textcolor{black}{\emph{MAP}} & \textcolor{black}{\emph{N@10}} & \textcolor{black}{\emph{Dim.}}\tabularnewline
\hline 
\textcolor{black}{500D-BOW$\rightarrow$ Tags} & \textcolor{black}{0.294 (+8.1)} & \textcolor{black}{0.458 (+4.3)} & \textcolor{black}{200} & \textcolor{black}{0.272} & \textcolor{black}{0.439} & \textcolor{black}{500}\tabularnewline
\textcolor{black}{64D-CH$\rightarrow$ Tags } & \textcolor{black}{0.279 (+2.6)} & \textcolor{black}{0.449 (+1.1)} & \textcolor{black}{200} & \textcolor{black}{0.272} & \textcolor{black}{0.444} & \textcolor{black}{64}\tabularnewline
\textcolor{black}{1134D-Visual$\rightarrow$Tags} & \textcolor{black}{0.374 (+6.6)} & \textcolor{black}{0.531 (+2.7)} & \textcolor{black}{200} & \textcolor{black}{0.351} & \textcolor{black}{0.517} & \textcolor{black}{1134}\tabularnewline
\textcolor{black}{1134D-Visual$\rightarrow$(Visual \& Tags)} & \textcolor{black}{0.386 (+10.0)} & \textcolor{black}{0.539 (+4.3)} & \textcolor{black}{200} & \textcolor{black}{0.351} & \textcolor{black}{0.517} & \textcolor{black}{1134}\tabularnewline
\textcolor{black}{(64D-CH \& Tags)$\rightarrow$ Tags} & \textcolor{black}{0.420 (+25.0)} & \textcolor{black}{0.575 (+11.0)} & \textcolor{black}{200} & \textcolor{black}{0.336} & \textcolor{black}{0.518} & \textcolor{black}{1064}\tabularnewline
\hline 
\end{tabular}
\par\end{centering}

\protect\caption{Image retrieval results on NUS-WIDE. A retrieved image is considered
relevant if it shares at least one concept with the query. Each test
query is matched against other test images\emph{. }\textcolor{black}{(Features}\textcolor{black}{\emph{$\rightarrow$
}}\textcolor{black}{Aux. Tasks) means auxiliary tasks are used to
fine-tune the model given the features. 64D-CH means the 64 bin color
histogram, 1134D-Visual means all 6 visual features combined.} N@10
= NDCG evaluated at top 10 results. The symbol $\uparrow$ denotes
increase in performance compared to the baseline. \label{tab:image-retrieval}}
\end{table*}

Table~\ref{tab:image-retrieval} reports the results. It is clear
that multiple visual representations are needed for good content-based
retrieval quality. For baseline (normalized feature concatenation),
the MAP improves by 29\% from 0.272 with BOW representation to 0.351
with all visual features combined. Similarly, our method improves
by 27\% from 0.294 with BOW alone (\textcolor{black}{row: 500D-BOW$\rightarrow$
Tags) to 0.374 with 6 visual feature types (row: 1134D-Visual$\rightarrow$Tags).}
Textual tags are particularly useful to boost up the performance,
perhaps because tags are closer to concepts in term of abstraction
level. We believe that this is reflected in our deep model with tagging
as an auxiliary task, where a consistent improvement over the baseline
is achieved, regardless of visual representations. We hypothesize
that a representation that can predict the tags well is likely to
be semantically closer to the concept.

When tags are available at the retrieval stage, there is a large boost
in performance for both the baseline and our method (see the difference
between row \textcolor{black}{64D-CH$\rightarrow$ Tags and row (64D-CH
\& Tags)$\rightarrow$ Tags), reconfirming that multimodal fusion
is required. The difference between our method and the baseline is
that the improvement gap is wider, e.g., 50.5\% improvement with our
method versus 23.5\% with the baseline.}

Interestingly, adding visual reconstruction as auxiliary tasks also
improves the retrieval performance. More specifically, the MAP increases
from 0.374 with tagging alone (row: \textcolor{black}{1134D-Visual$\rightarrow$Tags})
to 0.386 with both tagging and visual reconstruction (row: \textcolor{black}{1134D-Visual$\rightarrow$(Visual
\& Tags)}). This is surprising because visual information is low-level
compared to tags and concepts. This support a stronger hypothesis
that a representation predictive of many auxiliary tasks could be
suitable for retrieval. 

We note in passing that our deep architecture can produce a more compact
representation than the typical multimodal representation. For example,
visual features and tags combined generate 2,034 dimensions, an order
of magnitude larger than the size of the deep representation (200).

\subsection{Multilabel learning \label{sub:Multilabel-learning}}

\begin{figure}
\begin{centering}
\begin{tabular}{ccc}
\includegraphics[width=0.3\textwidth,height=0.25\textwidth]{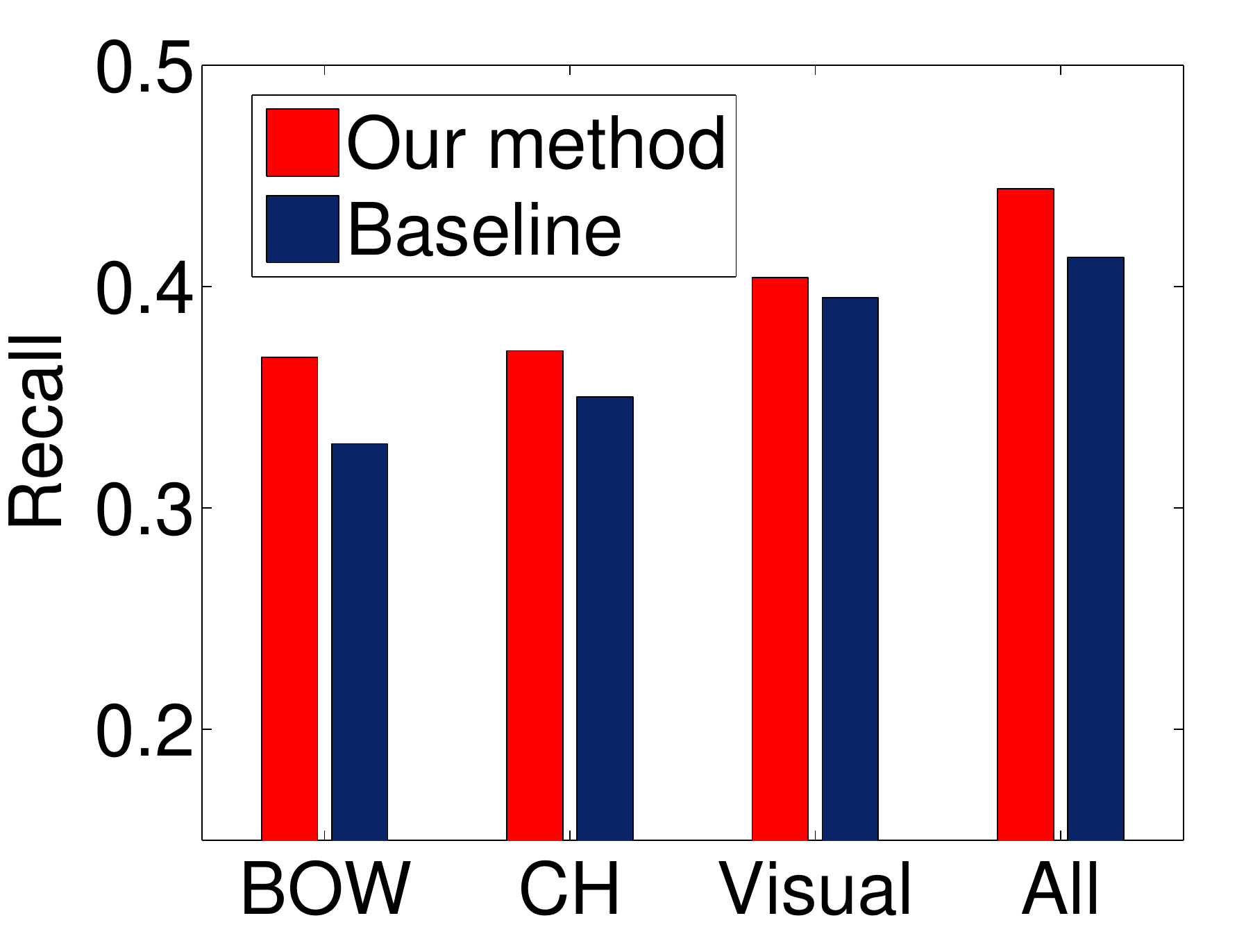} & \includegraphics[width=0.3\textwidth,height=0.25\textwidth]{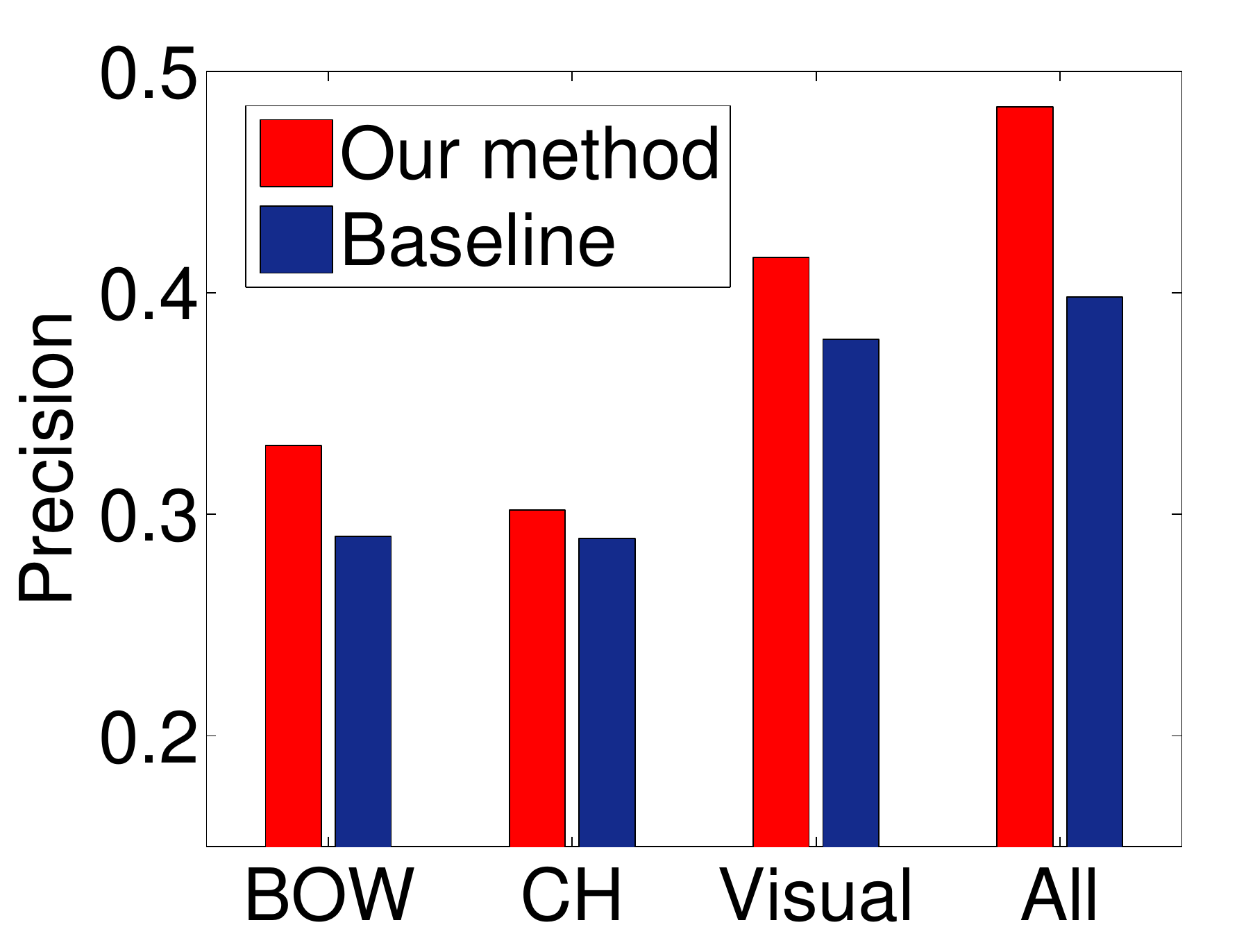} & \includegraphics[width=0.3\textwidth,height=0.25\textwidth]{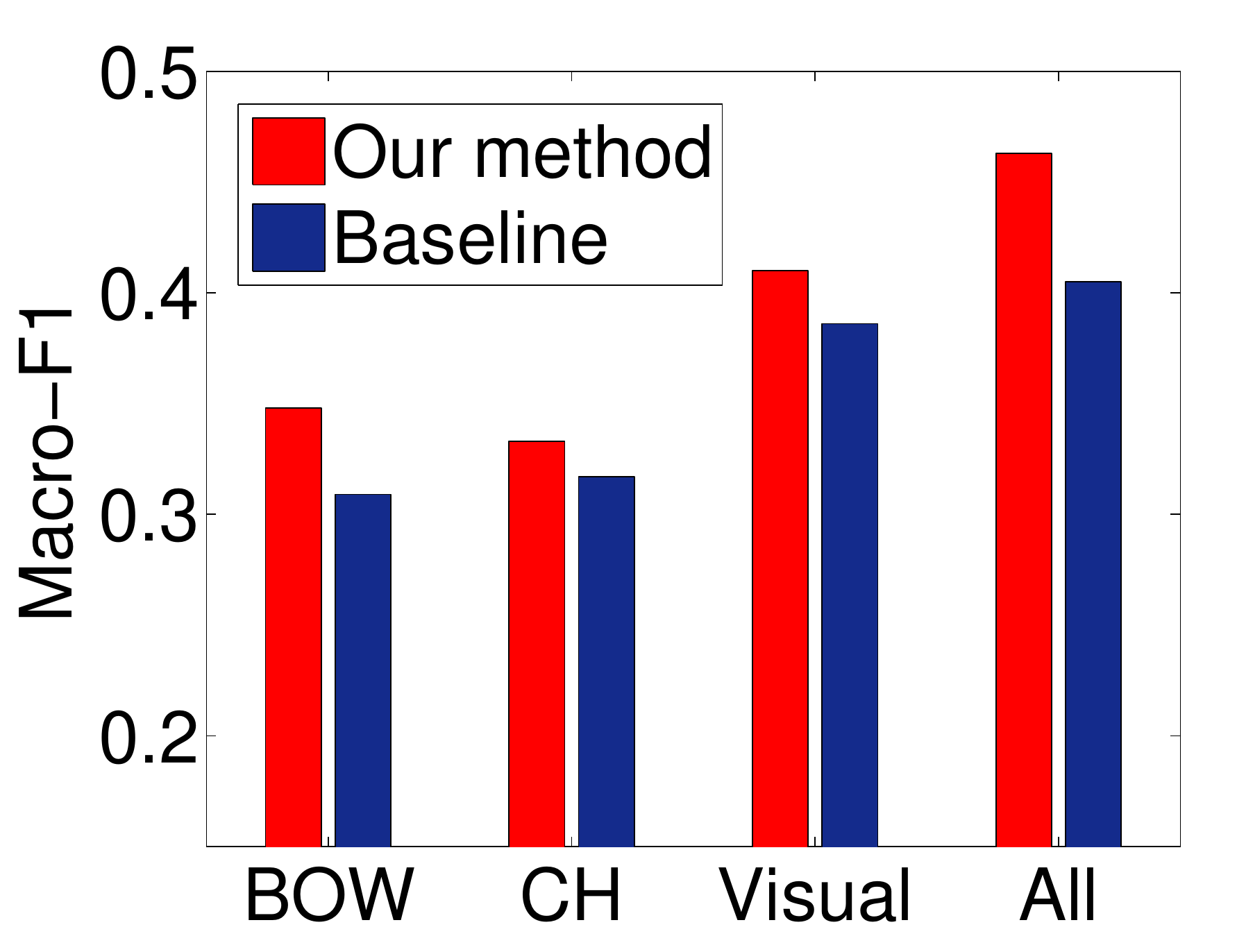}\tabularnewline
(a) Recall & (b) Precision & (c) Macro-F1\tabularnewline
\end{tabular}
\par\end{centering}

\protect\caption{Multiple concept labeling results. Tag ranking is used as auxiliary
task. Macro-F1 is the mean F-measure over all labels. BOW = 500D Bag-of-words
(counts), CH = 64D Color histogram (real-valued), Visual = all visual
views combined (counts \& real-valued), All = visual and tags (counts,
real-valued \& binary).\label{fig:Multiple-concept-labeling}}

\end{figure}

In this problem, we aim to predict high-level concepts from images.
There are $81$ concepts in total but each image is typically assigned
to $2$ concepts. This is cast as a multilabel prediction problem
(Sec.~\ref{sub:Structured-outputs}). To create a baseline, we use
a $k$-nearest neighbor classifier \cite{zhang2007ml}. First we
normalize multityped visual features as in Section~\ref{sub:Image-Retrieval}.
Then for the prediction, we retrieve top $30$ similar training images
for each test image and estimate label probabilities. Labels that
have higher probability than the estimated threshold will be selected.
Whenever possible, we use tag ranking as an auxiliary task in addition
to the main concept labeling task during training.

Figs.~\ref{fig:Multiple-concept-labeling}(a,b,c) plot the prediction
results on the test set under three performance metrics: recall, precision
and macro-F1. The results largely reflect the findings in the retrieval
experiments: (i) multiviews work better than single view, (ii) tags
are informative and thus should be exploited sensibly, (iii) the deep
architecture outperforms shallow one. In particular, fusing all visual
views leads to 21.8\% (baseline) and 23.1\% (deep) improvements in
macro-F1 over the single view of color histogram. When tags are integrated
as inputs, the improvement is higher: 39.0\% using our deep architecture,
and 27.8\% for the baseline. Our macro-F1 is 12.6\% better than the
baseline on BOW alone and 14.3\% better on the fusion of all types
and modalities.

\section{Conclusion \label{sec:Conclusion}}

We have introduced a deep architecture and a learning procedure to
discover joint homogeneous representation from multityped multimedia
objects. We have also presented the concept of multityped tasks which
deviate from the common single-type multitask setting. The deep learning
procedure has two phases, the unsupervised and supervised. The \emph{unsupervised
phase} starts from the bottom layer, where type-specific features
are modeled using separate RBMs, creating mid-level representations.
All these representations are then aggregated into a second level
RBM. In the \emph{supervised phase}, the two layers are then fused
into a deep neural network whose top layer is connected to one or
more tasks of interest. The whole network is then discriminatively
trained to obtain more predictive representation, starting from the
parameters estimated in the unsupervised phase. In summary, our architecture
seamlessly supports compact feature discovery from multimodal, multityped
data under multityped multitask settings. Its capacity has been demonstrated
on the tasks of image retrieval and multiple concept prediction showing
promising results.

It should be emphasized that the proposed architecture is modular.
Other types, more complex objects and tasks can be integrated easily.
Word counts, for example, can be modeled by replicated softmax \cite{salakhutdinov2009replicated},
an extension of the multinomial model (see also Eq.~(\ref{eq:multinomial})).
This paper is limited to a 3-layer deep architecture, but it can be
extended straightforwardly with more layers in each step.

%\section*{References}

\bibliographystyle{plain}

\end{document}